\documentclass[a4paper, 10pt, conference]{ieeeconf}      

\IEEEoverridecommandlockouts                              

\usepackage{amsmath} 

\usepackage{amsthm}
\usepackage{amsfonts}

\newtheorem{thm}{Theorem}

\usepackage{graphics} 
\usepackage{epsfig} 
\usepackage{epstopdf}
\usepackage{multirow}

\title{\LARGE \bf
Increased Mobility in Presence of Multiple Contacts - Identifying Contact Configurations that Enable Arbitrary Acceleration of CoM
}

\author{Milutin Nikoli\'c, Branislav Borovac, Mirko Rakovi\'c and Milica \v{Z}igi\'c
\thanks{Milutin Nikoli\'c, Branislav Borovac and Mirko Rakovi\'c are with the Faculty of Technical Sciences, University of Novi Sad, Novi Sad, 21000, Serbia {\tt\small \{milutinn, borovac, rakovicm\}@uns.ac.rs}.}
\thanks{Milica \v{Z}igi\'c is with the Faculty Of Sciences, University of Novi Sad, Novi Sad, 21000, Serbia {\tt\small mzigic@dmi.uns.ac.rs}.}
}

\begin{document}

\maketitle
\thispagestyle{empty}
\pagestyle{empty}

\begin{abstract}

Planning of any motion starts by planning the trajectory of the CoM. It is of the highest importance to ensure that the robot will be able to perform planned trajectory. With increasing capabilities of the  humanoid robots, the case when contacts are spatially distributed should be considered. In this paper, it is shown that there are some contact configurations in which any acceleration of the center of mass (CoM) is feasible. The procedure for identifying such a configurations is presented, as well as its physical meaning. On the other hand, for the configurations in which the constraint on CoM movement exists, it will be shown how to find that linear constraint, which defines the space of feasible motion. The proposed algorithm has a low complexity and to speed up the procedure even further, it will be shown that the whole procedure needs to be run only once when contact configuration changes. As the CoM moves,  the new constraints can be calculated from the initial one, thus yielding significant computation speedup. The methods are illustrated in two simulated scenarios. 

\end{abstract}

\section{INTRODUCTION}
The main indicator of dynamical balance during bipedal walking, $ZMP$ was introduced by M. Vukobratovi\'c and his closest associates \cite{vukobratovic69,vukobratovic04}. The $ZMP$ notion considers only the existence of full contact between the feet and the ground. It is assumed that the friction is high and that sliding does not occur. When the $ZMP$ is inside the convex hull of the support area (but not on its edges), the humanoid is dynamically balanced and the robot will not start falling  by rotating about the edge of the support area. The $ZMP$ notion in its original form can be used only for robots walking on flat horizontal surfaces. 
 
 Several authors have tried to generalize the \emph{ZMP} notion. Harada et al. \cite{harada06tro} proposed generalized $ZMP$ ($GZMP$) which is applicable for the cases when a robot is touching the environment with its hand. In order for the motion to be feasible, the $GZMP$ needs to fall within the region on the ground surface, defined on the basis of the 3D convex hull of all the contact points and mass of the robot. The two major assumptions used: the humanoid robot is standing/walking on a \emph{flat} floor and friction forces are \emph{small enough}, make this approach less general. The same authors analyzed the influence of additional contact (i.e., grasping the environment) on the position of $ZMP$ \cite{harada04}, and exploited that contact in order to enable the robot to climb up the high step by holding the handlebar. In \cite{harada06icra} the authors have proposed a general method for checking dynamical balance with multiple spatially distributed contacts, but it is assumed that friction is high, so that sliding does not occur. 
 
In \cite{sentis10}, the  authors have addressed the issue of robot control in the presence of multiple contacts by employing a virtual linkage model. Based on it, internal and contact forces required to maintain stability of all contacts are calculated. In \cite{takao03} feasible solution of wrench (FSW) is introduced. Gravitational and inertial forces have to be counter-balanced by contact forces so that their sum has to be in the space of feasible contact wrenches. Also, the authors have addressed the issue of low friction contact by introducing another criterion which can determine whether planned motion is feasible. In one of the recent advances \cite{caron15rss}, the authors have employed the double descriptor method \cite{fukuda96} in order to calculate the gravito-inertial wrench cone. Although most general compared to all previous $ZMP$ generalizations, the algorithm has high computational complexity.  

A lot of work has been done in this field, and the reported motion feasibility conditions can cover a vast variety of cases. Nevertheless, none of the research tries to answer the question:``Is there a configuration of spatially distributed contacts in which arbitrary acceleration of CoM can be achieved?''. It is very important question to answer since that kind of contact configurations allow for increased mobility and are highly favorable. In such a cases, there is no need for ground surfaces to support the weight of the robot, but the distribution of the contacts and frictional forces enable robot to support its own weight.  Also, since there is no constraint on the CoM acceleration, during motion planning the robot can be approximated by the simple point mass. More complex cart-pole and inverted pendulum models are not needed, because their main purpose is to model the movement of the robot with bounded acceleration of CoM.

This paper will prove that contact configurations that enable arbitrary acceleration of CoM exist: the procedure on how to identify them will be presented as well. This is the main contribution of the paper. This paper also considers the other case, when movement of CoM is bounded. By closely considering the mechanics of the system, it will be shown how to efficiently calculate gravito-inertial wrench cones, which is the second contribution of the paper. The computation time is reduced by two orders of magnitude compared to the state of the art methods. This could be very beneficial for optimization-based control strategies \cite{kuindersma14, herzog15, nikolic15}, which are inherently very computationally expensive.

 The paper is organized as follows: in section \ref{sec:constraints} the procedure for identifying the contact configurations which allow arbitrary CoM acceleration is described, as well as the procedure for efficient calculation of  contact wrench cones. These are the main contributions of the paper. The results of the simulations are presented in section \ref{sec:simulation}. Conclusions are given in section \ref{sec:conclusion}.

\section{MOTION CONSTRAINTS}\label{sec:constraints}
\begin{figure}[btp]
	\centering
	\includegraphics[height=7cm]{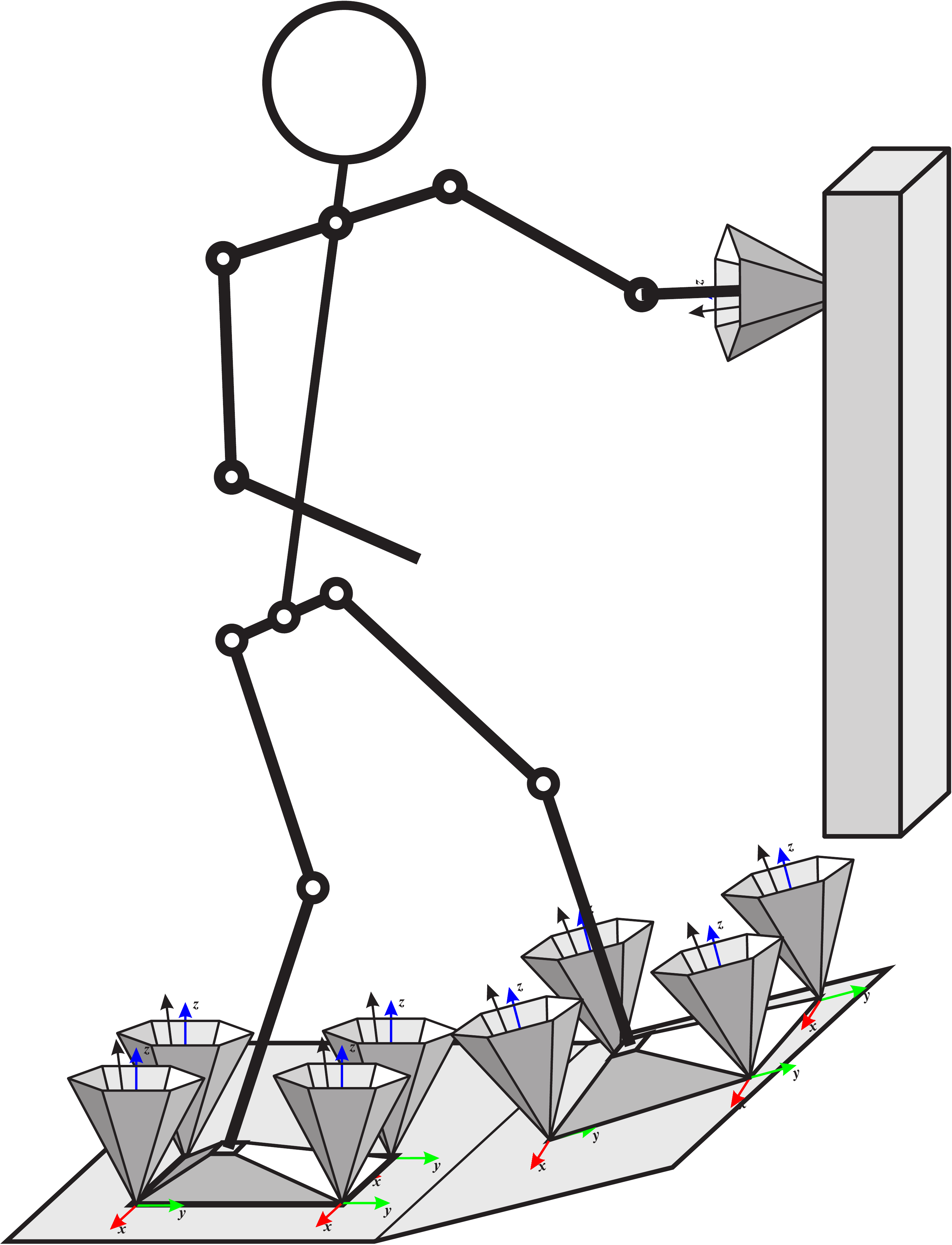}
	\caption{A system with multiple contacts. Contacts between the feet and the ground are considered planar while contact between the hand and the wall is considered as point contact}
	\label{fig:multiple_contacts}
\end{figure}

Let's consider the system shown in Fig. \ref{fig:multiple_contacts}. The robot is in contact with the environment with both of its feet and with one of its hands. Since the motion planning is mainly based on motion of the CoM, it is of highest importance to find what kind of motions of CoM does current contact configuration permit. Nature of the contacts induces constraints on contact forces, while their spatial disposition defines the space of feasible total wrench. For the system from the Fig. \ref{fig:multiple_contacts} acceleration of center of mass  and rate of change of angular momentum can be calculated using:
\begin{equation}
\label{eqn:kontakt:acmdes}
\begin{bmatrix}
m\mathbf{a} \\ \mathbf{\dot{L}}
\end{bmatrix}=
\begin{bmatrix}
\mathbf{F} \\ \mathbf{M}
\end{bmatrix} + \begin{bmatrix}
m\mathbf{g} \\ \mathbf{0}
\end{bmatrix}
\end{equation}
where $m$ is the mass of the system, $\mathbf{a}$ is the acceleration of the CoM, $\mathbf{L}$ is angular momentum, while $\mathbf{g}$ is gravitational acceleration. The total force acting on the system is denoted by $\mathbf{F}$, while the total torque that contact forces create about the CoM is denoted by $\mathbf{M}$. In order to ensure that the intended CoM motion (defined by  acceleration $\mathbf{a}$ and rate of change of angular momentum $\dot{\mathbf{L}}$)  is feasible there must exist a total wrench (which contains total force and total torque) such that \eqref{eqn:kontakt:acmdes} is fulfilled. Now the question of motion feasibility turns into a question of the existence of appropriate total forces and torques. 

\begin{figure}[btp]
	\centering
	\includegraphics[width=7cm]{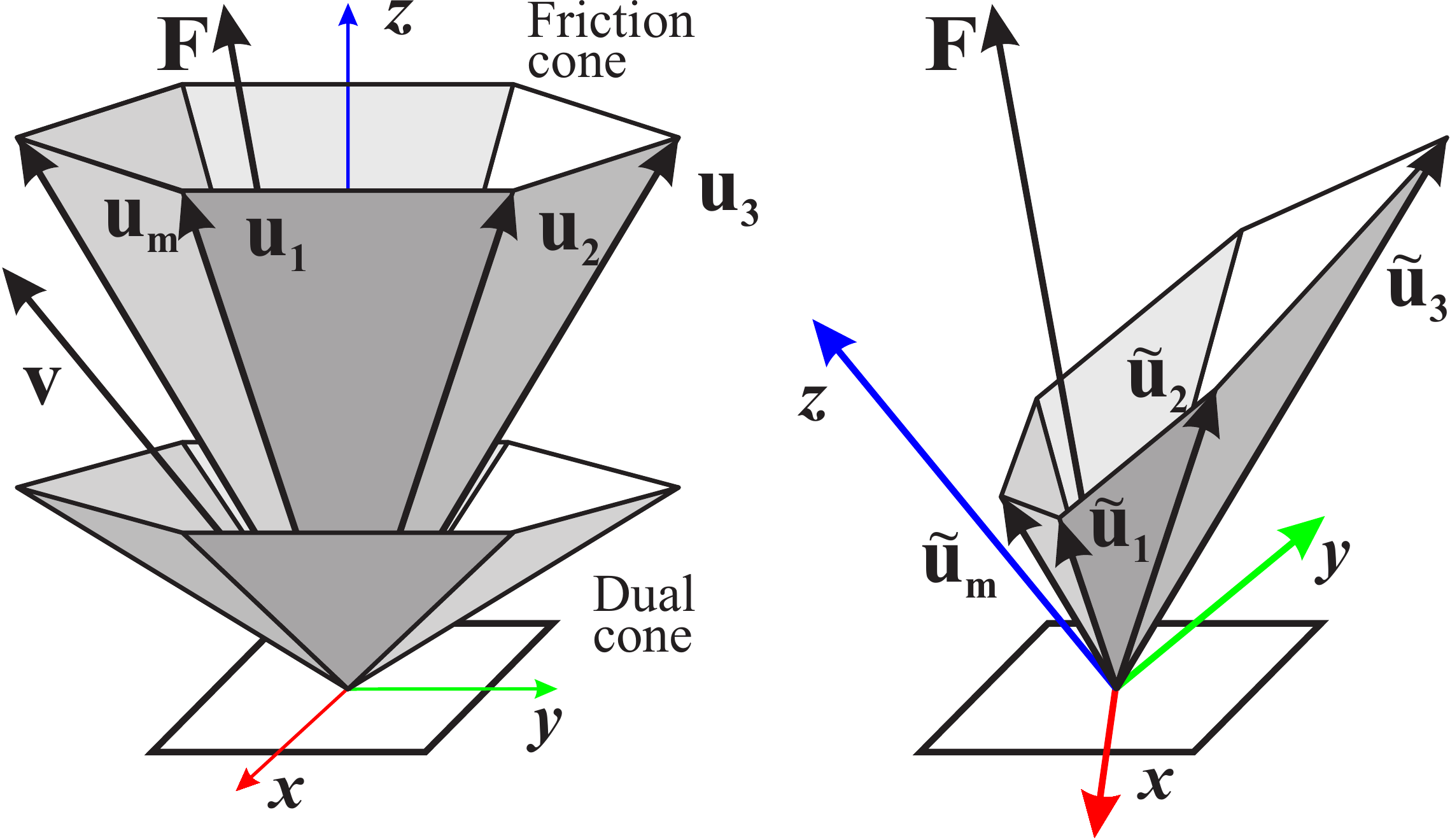}
	\caption{Friction cone with its dual is shown on the left. Modified friction cone with modified generating vectors is shown on the right. }
	\label{fig:dual_modified}
\end{figure}

As previously stated, the nature of the contact introduces the constraints on contact forces. In this paper only unilateral point contacts with Columb friction will be considered. Surface contacts can be understood as multiple point contacts placed at the corners of the perimeter of the contact surface\cite{caron15icra}. The contact force (denoted by $\mathbf{F}^{C}$) at the point contact is bounded, so it needs to be inside the corresponding friction cone which can be approximated by a $m$-sided pyramid (Fig. \ref{fig:dual_modified}). Representation of friction cone in the span form is:
\begin{equation}
\label{eqn:kontakt:Fspan}
\mathbf{F}^{C}=\alpha^1\mathbf{u}_1+\alpha^2\mathbf{u}_2+\dots+\alpha^m\mathbf{u}_m=\mathbf{U}\pmb{\alpha}, \;\; \alpha^i\geq 0
\end{equation}
where:
\begin{eqnarray}
\label{eqn:kontakt:ui}
\mathbf{u}_i=\begin{bmatrix}
\mu\cos\frac{2\pi (i-1/2) }{m}\\
\mu\sin\frac{2\pi (i-1/2) }{m}\\
1
\end{bmatrix}, \mathbf{U}=\begin{bmatrix}
\mathbf{u}_1 & 	\mathbf{u}_2 & \dots  & \mathbf{u}_m 
\end{bmatrix}
\end{eqnarray}
and $\pmb{\alpha}=\left[\alpha^1 \; \alpha^2 \;\dots\;\alpha^m\right]^T$. Matrix $\mathbf{U}$ represents generating matrix, which multiplied by vector $\pmb{\alpha}$ generates the space of feasible contact force. Equations \eqref{eqn:kontakt:Fspan} and \eqref{eqn:kontakt:ui} represent the constraints written in the local coordinate frame of the contact, where the $z$-axis represents normal at the surface of the point contact. In order for it to be expressed in the  global coordinate frame, the  matrix $\mathbf{U}$ needs to be pre-multiplied with an appropriate rotation matrix. 

For the case when multiple contacts are present (Fig. \ref{fig:multiple_contacts}) total wrench acting on the robot has the form:
\begin{eqnarray}
\label{eqn:kontakt:Ftotal}
&\begin{bmatrix}
\mathbf{F}\\
\mathbf{M}
\end{bmatrix}=\sum_{i=1}^n\begin{bmatrix}
\mathbf{R}_i\mathbf{F}^{C}_i\\ \left[\mathbf{r}_i\right]_\times\mathbf{R}_i\mathbf{F}^{C}_i
\end{bmatrix}=\\=&\begin{bmatrix}
\mathbf{R}_1\mathbf{U}_1 & \dots & \mathbf{R}_n\mathbf{U}_n
\\\left[\mathbf{r}_1\right]_\times\mathbf{R}_1\mathbf{U}_1 & \dots & \left[\mathbf{r}_n\right]_\times\mathbf{R}_n\mathbf{U}_n\end{bmatrix}
\begin{bmatrix}
\pmb{\alpha}_1 \\ \vdots \\ \pmb{\alpha}_n
\end{bmatrix}=\begin{bmatrix}
\pmb{\Upsilon}\\
\pmb{\Gamma}
\end{bmatrix}\mathbf{A}\nonumber
\end{eqnarray}
where $n$ is the number of separate contact points, $\mathbf{R}_i$ is the rotation matrix and $\mathbf{U}_i$ is the generating matrix for the $i$-th contact. Vectors $\mathbf{r}_{1}$ to $\mathbf{r}_{n}$ represent positions of contact points relative to CoM of the system.  $\left[ \bullet \right]_\times$ is a skew-symmetric cross product operator matrix. All matrices $\mathbf{U}_i$ can differ in terms of the friction coefficient $\mu$ and the number of sides of friction cone $m$, but for easier derivation, without loss of generality, only the case when all generating matrices have the same number of sides will be considered. Matrix $\pmb{\Upsilon}$ represents the matrix where all generating matrices are stacked horizontally, $\pmb{\Gamma}$ represents the matrix where all cross products of position vectors and generating matrices are stacked horizontally and $\mathbf{A}$ represents the vector of $\pmb{\alpha}_i$-s for all $n$ contacts. In the rest of the paper $i$-th columns of matrices  $\pmb{\Upsilon}$ and $\pmb{\Gamma}$ will be denoted as  $\pmb{\upsilon}_i$ and $\pmb{\gamma}_i$ respectively. 

The key problem here is to find the space of feasible forces acting on CoM $\mathbf{F}$ for the given contact configuration. Space of forces acting on CoM directly influences the space of feasible accelerations of CoM. For such a purpose the Farkas-Minkowski's theorem \cite{Farkas02}, in form taken from \cite{gale51}, will be employed: 
\begin{thm}[Farkas-Minkowski]
Let $\mathbf{A}$ be a real $m \times n$ matrix and $\mathbf{b}$ an $m$-dimensional real vector. Then, exactly one of the following two statements is true:
\begin{enumerate}
	\item There exists $\mathbf{x}\in\mathbb{R}^{n}$ such that $\mathbf{Ax} = \mathbf{b}$ and $\mathbf{x} \geq 0$.
	\item There exists $\mathbf{y} \in \mathbb{R}^{m}$ such that $\mathbf{y}^{\mathtt{T}}\mathbf{A} \geq 0$ and $\mathbf{y}^{\mathtt{T}} \mathbf{b} < 0.$	
\end{enumerate}
\end{thm}
 When that theorem is applied to case \eqref{eqn:kontakt:Ftotal} we get:
\begin{thm}
	\label{thm:nikolic}
	For the robot system in contact with the environment, with a matrix of generating vectors  $\pmb{\Upsilon}$, the total force acting on robot $\mathbf{F}$ is feasible if and only if:		
	\begin{itemize}
		\item For every vector $\mathbf{v}\in\mathbb{R}^3$ such that $\mathbf{v}^{\mathtt{T}}\pmb{\Upsilon} \geq 0$ it holds that $\mathbf{v}^\mathtt{T}\mathbf{F}\geq 0$	
	\end{itemize}
\end{thm}
Because this theorem might seem very abstract and hard to apply, it will be split in two cases in order to show its full potential. In the first case, we will consider the singular case where only zero-vector ($\mathbf{v}=\mathbf{0}$) satisfies $\mathbf{v}^{\mathtt{T}}\pmb{\Upsilon} \geq 0$. The second case considered will be the case where $\mathbf{v}\neq \mathbf{0}$  with the property $\mathbf{v}^{\mathtt{T}}\pmb{\Upsilon} \geq 0$ exists and in that case we will try to find the space of forces $\mathbf{F}$, where $\mathbf{v}^\mathtt{T}\mathbf{F}\geq 0$.

\subsection{First Case}\label{sec:first_case}
If we take closer a look at theorem \ref{thm:nikolic}, it can be concluded that if  $\mathbf{v}=\mathbf{0}$, then $\mathbf{v}^\mathtt{T}\mathbf{F}\geq 0$ no matter  the value of the total force $\mathbf{F}$. As a result it can be noted that this condition does not depend on total force at all, but only on matrix $\pmb{\Upsilon}$, which describes contact configuration. In other words, \emph{there are some contact configurations in which arbitrary total force exists\footnote{Of course, on a real robot, the force that can be achieved is limited by available joint driving torques}}. 

\begin{figure}[btp]
	\centering
	\includegraphics[height=6cm]{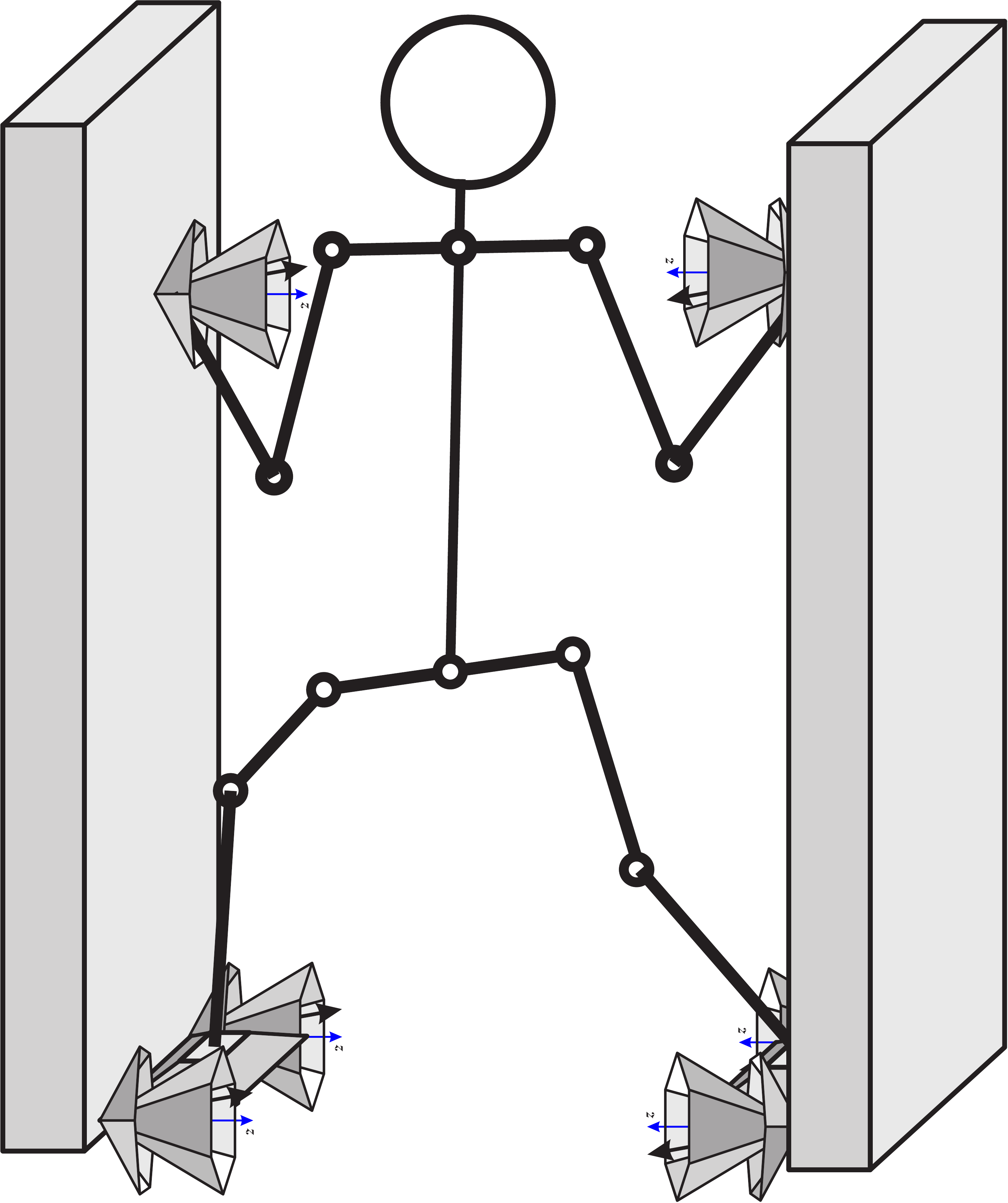}
	\caption{Robot pushing against two vertical walls. Duals of all six friction cones intersect only at $\mathbf{0}$, thus any acceleration of CoM is feasible}
	\label{fig:no_intersection}
\end{figure}

If for a vector $\mathbf{v}$ holds that $\mathbf{v}^{\mathtt{T}}\mathbf{R}_i\mathbf{U}_i\geq 0$ it means that vector $\mathbf{v}$ must lie in the dual cone\footnote{Dual cone of cone $C$ in three dimensional space is defined as $C^*=\left\{\mathbf{y} \in \mathbb{R}^3 : \mathbf{y}^T\mathbf{x}\geq0,\forall\mathbf{x} \in C\right\}$. In words, dual cone of a cone is a set of vectors whose dot product with all vectors from the original cone is positive.} of the friction cone of $i$-th contact. After looking at the structure of matrix $\pmb{\Upsilon}$ \eqref{eqn:kontakt:Ftotal} , it can be concluded together with theorem \ref{thm:nikolic} that, when the intersection of  dual cones of all contacts is  $\left\{\mathbf{0}\right\}$, the arbitrary desired total force exists, as well as the arbitrary acceleration of the CoM. An illustration of such a case is given in Fig. \ref{fig:no_intersection}, from which it can be clearly noted that there is no vector $\mathbf{v}\neq\mathbf{0}$ which would lie in all the friction cone duals.

In order to check if the dual cones intersect, we can solve the following linear optimization problem:
 \begin{eqnarray}
 \label{eqn:kontakt:program_v}
 \text{minimize:} & s\nonumber\\
 \text{such that:} & \pmb{\Upsilon}^{\mathtt{T}}\mathbf{v} + s \geq 0\\ 
 & s \geq -1\nonumber
 \end{eqnarray}
The solution of this problem is slack variable $s$ and vector $\mathbf{v}$, and since it is a convex problem it is guaranteed that a global minimum will be found. The last constraint ($s\geq -1$) is added so that the solution does not drift to $-\infty$. The case when the result of this linear program (LP) is  $s = 0$ and $\mathbf{v}=\mathbf{0}$, according to theorem \ref{thm:nikolic}, it means that all total forces acting on the robot $\mathbf{F}$ are feasible. Hence, arbitrary acceleration of the center of mass is feasible.  

On the other hand, when $s < 0$, the vector $\mathbf{v}$ that satisfies $\mathbf{v}^{\mathtt{T}}\pmb{\Upsilon} \geq 0$ is found. So in order to find the feasible space of total force $\mathbf{F}$ the second case must be considered. 

\subsection{Second case}\label{sec:second_case}
\begin{figure}[btp]
	\centering
	\includegraphics[height=6cm]{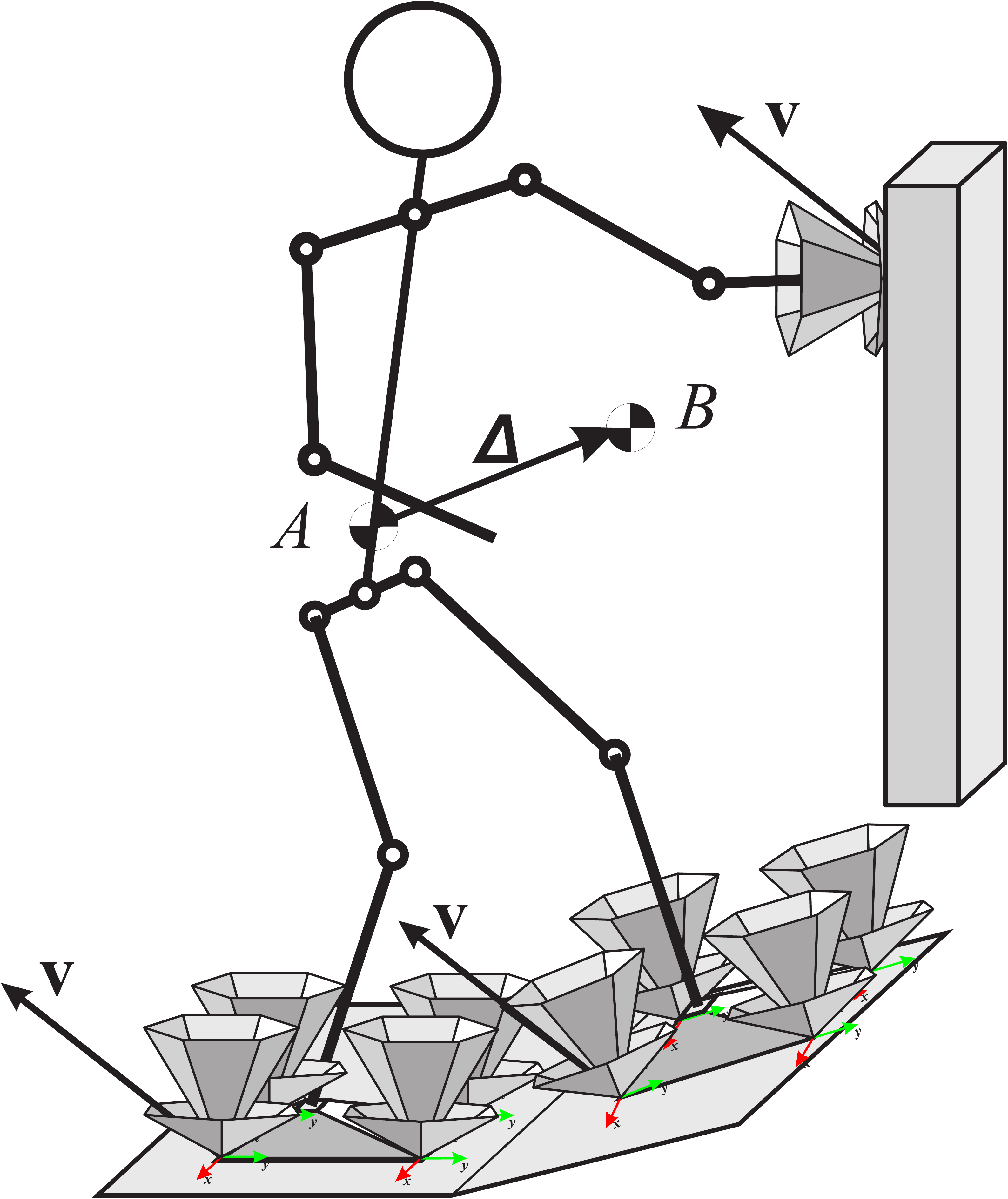}%
	\caption{Case where duals of friction cones intersect. Vector $\mathbf{v}$ that lies in duals of all friction cones is depicted.}
	\label{fig:second_case}
\end{figure}

When the vector $\mathbf{v}\neq0$, that lies in the intersection of the dual cones of all contacts, is found (Fig. \ref{fig:second_case}), it is possible to calculate the rotation matrix $\mathbf{R_v}$, which rotates the global coordinate frame so that the direction of a new $z$-axis aligns with the vector $\mathbf{v}$. An example of such a rotation is depicted in Fig. \ref{fig:dual_modified}. After rotation the total wrench acting on the body would be:
\begin{equation}
\label{eqn:FtotalMtotal}
\begin{bmatrix}
\mathbf{F^v}\\\mathbf{M^v}
\end{bmatrix}=\begin{bmatrix}
\mathbf{R_v}\pmb{\Upsilon}\\ \mathbf{R_v}\pmb{\Gamma}
\end{bmatrix}\mathbf{A}
\end{equation}
Since $z$-axis is aligned with $\mathbf{v}$ and $\mathbf{v}^\mathtt{T}\pmb{\Upsilon}\geq 0  $, all elements in the third row of matrix  $\mathbf{R_v}\pmb{\Upsilon}$ are greater than zero. Thus, columns of matrices $\mathbf{R_v}\pmb{\Upsilon}$ and $\mathbf{R_v}\pmb{\Gamma}$ would be normalized in order to obtain modified generating matrices:
\begin{eqnarray}
\tilde{\pmb{\upsilon}}_i =\frac{|\mathbf{v}|}{\mathbf{v}\bullet\pmb{\upsilon}_i}\mathbf{R_v}\pmb{\upsilon}_i, \;\; \tilde{\pmb{\Upsilon}}=\left[\tilde{\pmb{\upsilon}}_i\right], \;i=1\dots n\times m\\
\tilde{\pmb{\gamma}}_i =\frac{|\mathbf{v}|}{\mathbf{v}\bullet\pmb{\upsilon}_i}\mathbf{R_v}\pmb{\gamma}_i, \;\; \tilde{\pmb{\Gamma}}=\left[\tilde{\pmb{\gamma}}_i\right], \;i=1\dots n\times m
\end{eqnarray}
The friction cone and its generating vectors after modification are depicted in Fig. \ref{fig:dual_modified}. It is clear that both the original friction cone and the modified friction cone generate the same space, but now, friction cones of all contacts would be expressed in the same coordinate frame. To achieve this, equation \eqref{eqn:FtotalMtotal} would assume the following form:
\begin{equation}
\label{eqn:FtotalMtotalTilde}
\begin{bmatrix}
\mathbf{F^v}\\\mathbf{M^v}
\end{bmatrix}=\begin{bmatrix}
\tilde{\pmb{\Upsilon}}\\ \tilde{\pmb{\Gamma}}
\end{bmatrix}\tilde{\mathbf{A}}=\mathbf{P}\tilde{\mathbf{A}}
\end{equation}
 After modification, the third row of the matrix $\tilde{\pmb{\Upsilon}}$ would be a row of ones, and thus the force in $z$-direction of the new coordinate frame would be the sum of modified alphas ($F_z^{\mathbf{v}}=\sum_{i=1}^{n}\sum_{j=1}^{m}\tilde{\alpha}_i^j$). Now the third row from both sides of \eqref{eqn:FtotalMtotalTilde} would be removed and both sides would be divided by $F_z^{\mathbf{v}}$. As a result, we obtain:
 
 \begin{equation}
 \mathbf{p}_{-z}=\begin{bmatrix}
 F^{\mathbf{v}}_x/F^{\mathbf{v}}_z \\ F^{\mathbf{v}}_y/F^{\mathbf{v}}_z \\M^{\mathbf{v}}_x/F^{\mathbf{v}}_z\\M^{\mathbf{v}}_y/F^{\mathbf{v}}_z\\M^{\mathbf{v}}_z/F^{\mathbf{v}}_z
 \end{bmatrix}=\mathbf{p}_1\frac{\alpha_1^1}{F^{\mathbf{v}}_z}+\mathbf{p}_2\frac{\alpha_1^2}{F^{\mathbf{v}}_z}+\dots+\mathbf{p}_{n\times m}\frac{\alpha_n^m}{F^{\mathbf{v}}_z},
 \end{equation}
 where $\mathbf{p}_{-z}$ represents a 5D vector obtained from the contact wrench by removing normal force $F^{\mathbf{v}}_z$ and dividing the vector by $F^{\mathbf{v}}_z$, where $\mathbf{p}_i$ represents a point in 5D space which corresponds to i-th column of matrix $\mathbf{P}$ with the third row removed. Since $\sum_{i=1}^{n}\sum_{j=1}^{m}\frac{\tilde{\alpha}_i^j}{F^{\mathbf{v}}_z}=1$, $\tilde{\alpha}_i^j\geq0$ and $F^{\mathbf{v}}_z\ge0$ it can be concluded that the left-hand side is a convex combination of the points $\mathbf{p}_i$, thus the vector $\mathbf{p}_{-z}$ \emph{must lie within the convex hull of the points $\mathbf{p}_i, i=1\dots n\times m$}.
 
 The procedure for calculating the convex hull of points in hyperspace is well known and has low complexity. For the Quickhull algorithm implemented in Matlab, the average case complexity is  $O\left(n_{points} \log n_{points}\right)$\cite{barber96}, which reduces to $O\left(n m \log\max(n,m) \right)$ for the case considered here. As a result of the Quickhull the hyperplanes that bound the convex hull are obtained. Each of the hyperplanes introduces one inequality constraint that can be written in the form:
 \begin{equation}
 a_1 \frac{F^{\mathbf{v}}_x}{F^{\mathbf{v}}_z} +a_2 \frac{F^{\mathbf{v}}_y}{F^{\mathbf{v}}_z}+a_3 \frac{M^{\mathbf{v}}_x}{F^{\mathbf{v}}_z}+a_4 \frac{M^{\mathbf{v}}_y}{F^{\mathbf{v}}_z}+a_5 \frac{M^{\mathbf{v}}_z}{F^{\mathbf{v}}_z} \geq w
 \end{equation} 
 where $a_1\dots a_5$ and $w$ are parameters of the hyperplane. After multiplying both sides with $F^{\mathbf{v}}_z$ and rearranging the terms in the equation, it can be seen that one hyperplane from the convex hull introduces one inequality constraint on the wrench acting on CoM expressed in a rotated referent frame:
 \begin{equation}
 \begin{bmatrix}
 a_1 & a_2 & -w & a_3 & a_4 & a_5
 \end{bmatrix}\begin{bmatrix}
 \mathbf{F}^{\mathbf{v}} \\ \mathbf{M}^{\mathbf{v}}
 \end{bmatrix}\geq 0
 \end{equation}
 Each hyperplane introduces one constraint in the previous form, so after stacking all those constraints in one matrix and expressing everything in an original coordinate frame it can be obtained:
 \begin{equation}
 \label{eqn:kontakt:W}
 \mathbf{W}\begin{bmatrix}
 \mathbf{F} \\ \mathbf{M}
 \end{bmatrix}\geq 0
 \end{equation}
 where $\mathbf{W}$ represents wrench constraint matrix about CoM - \emph{WCM}. It is shown here how constraint on total wrench acting on the body can be constructed when duals of all friction cones intersect. By following the derivation from this chapter in reverse it can be easily proven that if condition \eqref{eqn:kontakt:W} is satisfied then $\mathbf{v}^\mathtt{T}\mathbf{F}=F_z^{\mathbf{v}}=\sum_{i=1}^{n}\sum_{j=1}^{m}\tilde{\alpha}_i^j\geq 0$. Together with intersecting duals, this is both  necessary and a sufficient condition for existence of feasible force according to theorem \ref{thm:nikolic}. 

 \subsection{Shifted WCM}
 As stated in the previous paragraph, WCM is calculated for the CoM of the system. Thus, the previously calculated WCM is no longer valid once the CoM changes its location. Since the robot moves, its CoM also shifts. Each time the CoM shifts the WCM could be recalculated, but by considering the mechanics it can be shown that it is possible to shift WCM, once it is computed, which makes the whole procedure even faster.
 
  If the WCM is calculated for point $A$ (current position of CoM) and the CoM shifts by vector $\pmb{\Delta}$ to some point $B$ (Fig. \ref{fig:second_case}), the wrench acting on the new position of CoM can be calculated for point A:
 \begin{equation}
	 \label{eqn:kontakt:FaMaFbMb}
	\begin{bmatrix}
		\mathbf{F}^A\\
		\mathbf{M}^A	
	\end{bmatrix}=\begin{bmatrix}
	\mathbf{I}_{3 \times 3} & \mathbf{0}_{3\times 3}\\ \left[\pmb{\Delta}\right]_{\times} & \mathbf{I}_{3 \times 3} ]
	\end{bmatrix}\begin{bmatrix}
	\mathbf{F}^B\\
	\mathbf{M}^B	
	\end{bmatrix}.
 \end{equation}
Since WCM is calculated for point A, wrench calculated for point A must fulfill inequality \eqref{eqn:kontakt:W}. So after combining that \eqref{eqn:kontakt:W} with \eqref{eqn:kontakt:FaMaFbMb} it is easy to obtain:
 \begin{eqnarray}
 \label{eqn:kontakt:Wdelta}
& \mathbf{W}^{\pmb{\Delta}}=\mathbf{W}\begin{bmatrix}
 \mathbf{I}_{3 \times 3} & \mathbf{0}_{3\times 3}\\ \left[\pmb{\Delta}\right]_{\times} & \mathbf{I}_{3 \times 3} ]
 \end{bmatrix}&\\
  \label{eqn:kontakt:WdeltaFbMb}
& \mathbf{W}^{\pmb{\Delta}}\begin{bmatrix}
 \mathbf{F}^B\\
 \mathbf{M}^B	
 \end{bmatrix}\geq 0&
 \end{eqnarray}
 where $\mathbf{W}^{\pmb{\Delta}}$ represents a shifted WCM. This shows, that WCM needs to be computed only once when the contact configuration changes. As CoM moves, as long as the contact configuration remains the same, the shifted WCM can be calculated by \eqref{eqn:kontakt:Wdelta}, which is a very fast operation to perform. Shifted WCM gives the same constraint \eqref{eqn:kontakt:WdeltaFbMb} on the total wrench as WCM \eqref{eqn:kontakt:W}, but saves the computation time by bypassing the procedure described in section \ref{sec:second_case}.

\section{SIMULATON RESULTS}\label{sec:simulation}
 \begin{figure*}[bthp]
 	\centering
 	\includegraphics[width = 18cm]{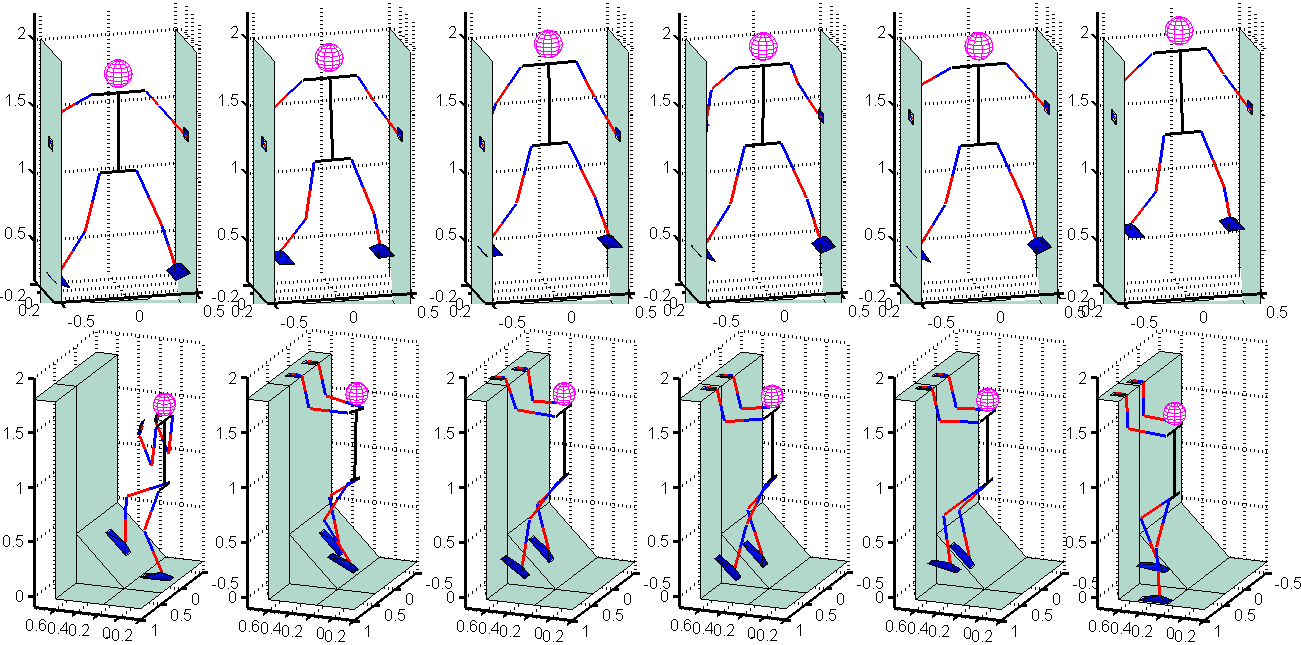}
 	\centering
 	\caption{Top row: A Robot climbing between the vertical walls. In diagrams 2 and 6 from the left - the robot is only in contact with the wall by his hands, while he repositions its feet. In diagram 4, edges of the feet are in contact with the wall while the robot repositions its hands. In diagrams 1, 3, and 5 the robot is in transition phase where feet and hands are in contact with the wall\protect\linebreak Bottom row: The robot moves sideways while hanging on the top of the wall. Six characteristic phases are shown. Contact configuration is described in Table \ref{tab:duration}}
 	\label{fig:simulation}
 \end{figure*}
In this section, the main results will be illustrated by two simulated scenarios. In the first case, the robot has to climb between two vertical walls as illustrated in Fig. \ref{fig:no_intersection}. In the second simulated case, the robot has to move sideways on the vertical wall while hanging by its hands at the top. The joint torques are calculated using a generalized task prioritization framework. More details about the framework can be found in \cite{nikolic13, nikolic15}. In both cases, all friction cones are approximated by 4-sided pyramids and $\mu=0.8$.

In the first simulated case only the hands and the edges of the feet are in contact with the wall. Most of the time only hands or only feet are in contact with the wall, except the short transition time between those two contact configurations. The illustrations of one and a half cycle of simulated motion are shown in the top row of Fig. \ref{fig:simulation}. The robot first repositions its feet and after that repositions its hands.  As the contact configuration changes the LP \eqref{eqn:kontakt:program_v} needs to be solved.  When only edges of the feet are in contact (diagram 4 from Fig. \ref{fig:simulation}), 4 point contacts are present  and the matrix $\pmb{\Upsilon}$  would be of size $3\times16$. The LP would be solved on average in $0.2ms$. On the other hand, when 8 contacts are present (hands are in contact with the wall, diagrams 2, and 6) it would take an average $0.35ms$ to solve the LP. During the transition phase (diagrams 1, 3, and 5), when all 12 contacts are present, the LP would take an average $0.55ms$. The procedure was implemented in Python together with Numpy and CVXOPT packages. Each time the LP is solved, as anticipated, the vector $\mathbf{v}$ degenerates to zero and $s$ would be equal to 0. According to the the first case (section \ref{sec:first_case}), we can see that  any total force $\mathbf{F}$ is feasible for contact configurations that appear in this scenario, implying that arbitrary acceleration of the CoM is also feasible. As a result, the robot is able to climb between the vertical walls, although there is no ground support. Let us point out once more that the only thing that could hinder the robot from climbing are limited joint torques since internal load needs to be introduced in order to maintain all contacts stable. When enough torque can be provided, the robot is able to climb the wall. 

\begin{table}[!b]
	\centering
	
	\caption{Average time taken to calculated WCM}
	\label{tab:duration}
	\begin{tabular}{p{0.3cm} p{0.7cm} p{0.7cm} p{0.5cm} p{0.5cm} ||p{0.95cm}||p{0.7cm} p{0.7cm}}	
		{}&  \multicolumn{5}{c}{Contact configuration} & \multicolumn{2}{c}{Calculation times} \\\hline
		phase & Right Foot & Left Foot 	& Right Hand & Left Hand & No. of contacts & WCM $[ms]$	& shifted WCM $[ms]$ \\\hline
		1   & Inclined    & Ground  	& -        	 & -         &		8	& 2.34 		& 0.038 \\
		2   & Inclined    & -    	    & Top        & Top       &		12		& 4.29	 		& 0.047 \\
		3   & Inclined    & Vertical    & Top        & -       	 &		10		& 2.75 		& 0.041 \\
		4   & -           & Vertical    & Top        & Top       &		10		& 3.55 		& 0.042 \\
		5   & Inclined    & -    		& Top        & Top       &		12		& 4.2 			& 0.046 \\
		6   & -           & Ground      & Top        &  Top  	 &		12		& 3.17 		& 0.038\\
		\hline
	\end{tabular}
	
\end{table}

In the second case considered, the contact configuration is changed substantially as the motion progresses. Hands can be only in contact with the horizontal top surface of the wall while the robot's feet can be in contact with the ground surface, with the surface angled at $45^\circ$ or with a vertical wall, but only using front edge of the foot. In Fig. \ref{fig:simulation} a total of 6 phases of the motion are shown. After establishing the contact between the hands and the top of the wall, in each phase one of the limbs is repositioned, while others remain stationary. The sequence of repositioning is: left foot, left hand, right hand and right foot. Each phase has a different spatial contact configuration. In all phases of the motion, the duals of the friction cones intersect and the WCM has to be calculated. Average calculation times of the WCM are shown in Table \ref{tab:duration}\footnote{Contact with the ground surface is denoted with 'Ground', contact with the surface angled ad $45^{\circ}$ is denoted with 'Inclined', contact with the top of the wall is denoted with 'Top' and contact with the vertical part of the wall (in this case only the front edge of the robot's foot is in contact with the wall).}. It can be seen that the calculation takes the longest during phases 2 and 5 when 12 point contacts exist and the contacts are distributed in 2 different planes. In phase 6, the number of contacts is also 12, but since in that case all the planes are parallel, there is no need to run the LP \eqref{eqn:kontakt:program_v}, so the procedure takes about $1ms$ less. As the number of contacts decreases (phases 1, 3 and 4, see Table \ref{tab:duration}) the execution time will decrease as well. It is important to note that  WCM needs to be computed only once when the configuration changes and the shifted WCM could be used afterward. That reduces the computation time by almost two orders of magnitude and enables this approach to be used for real-time control of the humanoid robots. During the simulation the WCM constraint was used to check the feasibility of the planned motion. Shifted WCM and \eqref{eqn:kontakt:WdeltaFbMb} were used in order to speed up the procedure without any influence on the results. At this stage the motion and contact sequence were planned manually.

\section{CONCLUSIONS AND OPEN QUESTIONS}\label{sec:conclusion}

In this paper an efficient procedure is presented for checking if current contact configuration allows realization of planned motion. It is shown that contact configurations which allow arbitrary acceleration of the CoM exist. In addition, the geometrical interpretation of this condition is given, stating that when all friction cones intersect only at $\mathbf{0}$ arbitrary acceleration of CoM is feasible. Such configurations enable advanced mobility and unbounded movement of CoM in any direction. In the simulation example, it is illustrated how such kind of configuration could be identified and exploited for climbing between vertical walls without having horizontal support surfaces, which bears the weight of the robot. 

For the case when acceleration of the CoM is bounded, the procedure is presented for calculating a wrench constraint matrix. Instead of recalculating WCM each time the CoM moves, the procedure for shifting WCM is created which enables for very expedient computation times. Actually, the computational times of WCM dropped almost 100 times. This result is significant, since it will enable this constraint to be used for the real-time control of humanoid robots

The focus of this paper was on checking if desired motion is possible, without considering  the joint torque limitations.  From the contact configuration perspective, some motion might seem to be feasible, but the robot might not be able to produce sufficient joint torques, which provide needed contact forces in order to maintain stability of all contacts and produce desired motion. In order to resolve this issue, apart from contact configuration, the robot's configuration together with joint torque limitations need to be closely studied.






\section*{ACKNOWLEDGMENT}
This work was funded by the Ministry of Science and Technological Development of the Republic of Serbia in parts under contracts TR35003 and III44008. 

\addtolength{\textheight}{-12cm}   

\bibliographystyle{IEEEtran}
\bibliography{IEEEabrv,Bibliography}

\end{document}